\pdfoutput=1

\documentclass[11pt]{article}

\usepackage[final]{acl}
\usepackage{mathtools} 
\usepackage{makecell}
\usepackage{times}
\usepackage{latexsym}
\usepackage{amsmath}   
\usepackage{amsfonts}  
\usepackage{booktabs}  

\usepackage[T1]{fontenc}

\usepackage[utf8]{inputenc}
\usepackage{amsmath}
\usepackage{amsthm}
\usepackage{enumitem}
\usepackage{multirow}
\usepackage{subcaption}
\usepackage{booktabs}

\usepackage{microtype}

\usepackage{inconsolata}

\usepackage{graphicx}

\usepackage{enumitem}
\usepackage{listings}  
\usepackage{pifont}

\title{Towards Advanced Mathematical Reasoning for LLMs \\ via  First-Order Logic Theorem Proving}

\author{Chuxue Cao$^{1}$, 
Mengze Li$^{1}$, 
Juntao Dai$^2$, 
Jinluan Yang$^3$, 
Zijian Zhao$^1$ \\
\textbf{Shengyu Zhang$^3$, 
Weijie Shi$^1$,
Chengzhong Liu$^1$, Sirui Han$^{1*\dag}$, Yike Guo$^{1}$}\thanks{Corresponding author. $^{\dag}$Project leader.} \\
$^1$Hong Kong University of Science and Technology \\
$^2$Peking University~~$^3$Zhejiang University \\
\texttt{ccaoai@connect.ust.hk}~~~~~\texttt{\{siruihan, yikeguo\}@ust.hk}
}

\begin{document}
\maketitle
\begin{abstract}

Large language models (LLMs) have shown promising first-order logic (FOL) reasoning capabilities with applications in various areas. However, their effectiveness in complex mathematical reasoning involving multi-step FOL deductions is still under-researched. While LLMs perform competitively on established mathematical reasoning benchmarks, they struggle with multi-step FOL tasks, as demonstrated by Deepseek-Prover-V2-7B's low accuracy (4.2\%) on our proposed theorem proving dataset. This issue arises from the limited exploration of diverse proof strategies and the potential for early reasoning mistakes to undermine entire proofs. To address these issues, we propose DREAM, a self-adaptive solution that enhances the Diversity and REAsonability of LLMs' generation strategies. DREAM incorporates an Axiom-Driven Strategy Diversification mechanism to promote varied strategic outcomes and a Sub-Proposition Error Feedback to help LLMs reflect on and correct their proofs. Our contributions include pioneering advancements in LLMs' mathematical reasoning through FOL theorem proving, introducing a novel inference stage solution that improves performance by 0.6\% to 6.4\%, and providing a curated dataset of 447 mathematical theorems in Lean 4 format for evaluation. Our code is available~\footnote{https://github.com/chuxuecao/dream-fol-prover}.

\end{abstract}

\begin{figure}[t]
\centering
  \includegraphics[width=\columnwidth]{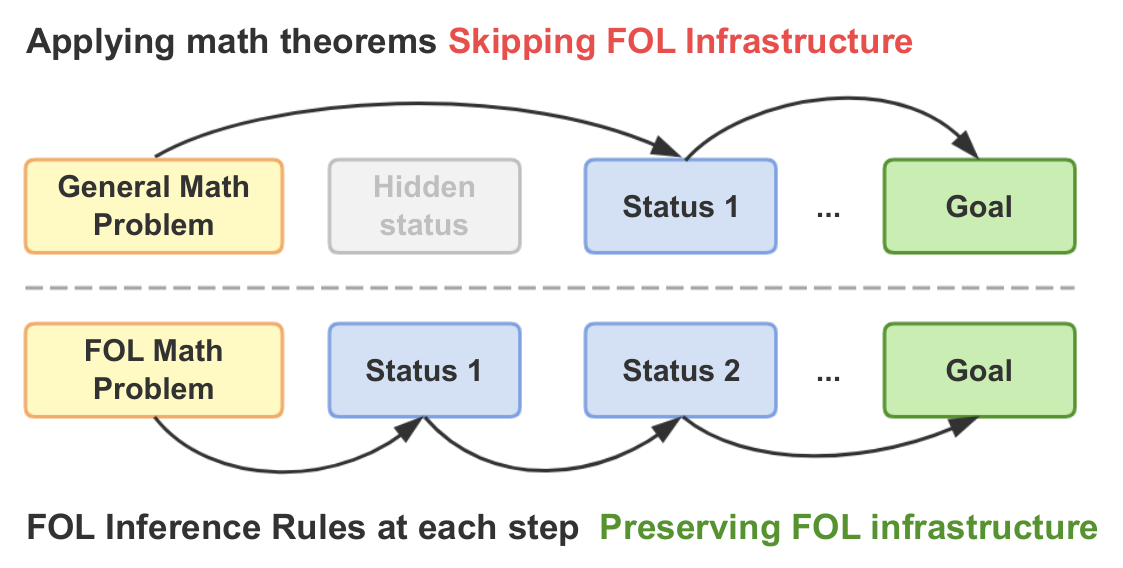}
  \caption{Distinction between First-Order Logic (FOL) and general math problems: FOL theorem proving requires strict stepwise adherence to FOL inference rules (e.g., universal instantiation, existential elimination), whereas general mathematical proving can utilize domain-specific mathematical theorems without explicitly referencing FOL infrastructure (including FOL rules and theorems). }
  \label{proposition-example}
\end{figure}

\section{Introduction}

Large language models (LLMs) have demonstrated emerging capabilities in first-order logic (FOL) reasoning~\citep{zhou2024rulearena}, with successful applications across legal precedent analysis~\citep{10353499} and logical fallacy detection~\citep{lalwani2025autoformalizingnaturallanguagefirstorder,ibragimov2025logical}. However, their efficiency in addressing complex mathematical reasoning tasks characterized by multi-step FOL deductions remains underexplored~\citep{CAO2021281}.

While contemporary LLMs attain competitive performance on established formal mathematical reasoning benchmarks such as miniF2F (formal Olympiad-level mathematics~\citep{zheng2022miniff}) and ProofNet (formal undergraduate-level mathematics~\citep{azerbayev2023proofnetautoformalizingformallyproving}), they reveal persistent deficiencies in mathematical reasoning with multi-step FOL deductions. Our controlled experiments demonstrate that DeepSeek-Prover-V2-7B~\citep{ren2025deepseekproverv2advancingformalmathematical} — despite comprehensive pretraining on Lean 4's formal mathematics corpus— achieves merely 4.2\% accuracy (pass@10) on our proposed FOL-based mathematical theorem proving tasks. This stark contrast between general and FOL-based mathematical reasoning capability, as shown in Figure \ref{proposition-example}, exposes limitations in current LLMs' capacity for handling nested quantifier interactions and negation propagation through extended deductive sequences~\citep{qi2025largelanguagemodelsmeet}.

For the FOL theorem proving problems, existing LLMs face two challenges: (i) \textit{The Adaptive Strategy Starvation Dilemma:} Unlike standard mathematical problems, where fixed solution methods often suffice, FOL proofs demand both tactical flexibility and strategic oversight. The high sensitivity of proof chains to initial assumptions requires exploring multiple proof strategies and maintaining logical consistency throughout the deduction process. But current training paradigms predominantly utilize fixed logical structures from proof assistant libraries (e.g., Lean 4's Mathlib)~\citep{lin2025goedelproverfrontiermodelopensource}, which encapsulate only a constrained subset of FOL applications, further preventing models from capturing the whole combinatorial space of potential logical constructions and reasoning patterns.  
While inference-stage solutions for general-domain math theorem proving are proposed to mitigate this issue~\citep{yang2023leandojo, pmlr-v235-zhao24h}, they overlook the specific features of FOL proving, restricting their efficiency. (ii) \textit{The Severe Cascading Error:} Within the reasoning chains for FOL theorem proving, early strategic errors can propagate through subsequent inferences, further undermining the entire proof, which can be defined as a cascading error~\citep{kovacs2013first, barwise1977introduction}. Compared with the modular error propagation seen in numerical calculations and code generation, the cascading error in FOL is more challenging due to the interdependence of logical steps and the lack of clear boundaries between errors. Thus, low-level error signals from a formal compiler are insufficient, as they fail to address the broader implications of flawed strategies.
 
To address the above challenges, we propose a novel inference stage solution that promotes the \textbf{D}iversity and \textbf{REA}sonability of LL\textbf{M}s' generation strategies, assisted by the detected errors across the entire proof, named \textbf{DREAM}. It includes two key designs:  \textbf{Axiom-Driven Strategy Diversification:} 
To avoid strategy starvation, we propose an axiom-driven strategy diversification mechanism based on a k-wise combinational axiom tree. This approach enables diverse strategy selection by focusing on different axioms, resulting in varied strategic outcomes. 
\textbf{Sub-Proposition Error Feedback:} To mitigate cascading errors, we propose a sub-proposition error feedback mechanism that aligns each error message with its corresponding sub-proposition using inline comments. This approach provides insights into the sub-propositions, encouraging LLMs to reflect on and revise their proof strategies thoroughly.

Our contributions are summarized as three-fold:

\begin{itemize}
    \item To the best of our knowledge, we are the first to advance LLMs' mathematical reasoning via FOL theorem proving, which especially requires LLMs to generate proof steps by strictly adhering to FOL rules and theorems. 

    \item We propose an inference stage solution through axiom-driven strategy diversification and sub-proposition error feedback mechanisms to enhance LLM's performance in this challenging FOL theorem proving task, achieving average gains from 0.6\% to 6.4\%.

    \item A carefully curated dataset is provided for extensive evaluation, containing 447 mathematical theorems from 10 categories within first-order logic written in Lean 4 format.
    
\end{itemize}

\section{Related Work}

\subsection{First-Order Logic Reasoning}

The interaction between FOL reasoning and LLMs manifests in two key directions: (i) leveraging FOL to enhance the faithfulness of LLM reasoning and (ii) evaluating LLM's long-chain deduction capabilities. Recent advancements illustrate this dual focus. For instance, LOGIC-LM~\citep{pan2023logiclmempoweringlargelanguage} and LINC~\citep{olausson2023linc} employ LLMs to translate natural language (NL) statements to formal FOL expressions, then utilize symbolic reasoning tools for verification and self-refinement, thereby grounding LLM outputs in rigorous logical frameworks. Concurrently, studies such as ~\citet{ryu2025divide}, \citet{qi2025largelanguagemodelsmeet}, and \citet{thatikonda2025assessingalignmentfolcloseness} propose algorithms for constructing high-quality FOL datasets and evaluating LLMs' multi-step reasoning capabilities. 

However, while these works mark significant progress, their datasets predominantly center on real-world scenarios (e.g., everyday life~\citep{han2024folionaturallanguagereasoning, tian-etal-2021-diagnosing, saparov2023language, tafjord-etal-2021-proofwriter, clark2020transformers}). A critical gap persists in formal FOL mathematical reasoning. Despite efforts to evaluate LLMs’ logical skills via NL-encoded FOL problems~\citep{ibragimov2025logicalskillslargelanguage}, their capacity to handle formal axiomatically defined systems (e.g., mathematical theorems, formal proof chains, or abstract logical relationships) remains underexplored. This omission limits understanding of LLMs’ ability to navigate domains where precision, symbolic rigor, and adherence to axiomatic structures are paramount. To fill this gap, we create a formal FOL reasoning dataset in the mathematical domain by utilizing the advanced FOL translation capabilities of LLMs. The detailed comparison between our datasets and previous datasets can be shown in Table~\ref{comp-datasets}.

\subsection{Formal Theorem Proving}

LLM-based theorem proving methods offer flexible control over problem complexity and diversity~\citep{johansson2023exploring, zhou2024refactorlearningextracttheorems, wu2022autoformalizationlargelanguagemodels, he2024bc, wan2024logicasker, xiong2023trigo}. Research in this area splits into two main approaches: complete proof generation and stepwise generation. For stepwise generation, models like BFS-Prover~\citep{xin2025bfsproverscalablebestfirsttree} and InternLM2.5-StepProver~\citep{wu2024internlm25stepproveradvancingautomatedtheorem} predict proof steps based on current status, while LeanDojo reduces hallucination through retrieval-based premise selection~\citep{yang2023leandojo}. In contrast, LEGO-prover and DTV focus on prompting LLMs for complete proofs~\citep{wang2023lego, zhou2024don}. Baldur enhances proof accuracy using error feedback~\citep{10.1145/3611643.3616243}, and~\citet{pmlr-v235-zhao24h} introduces a subgoal-based framework for LLMs. However, these methods have not optimized LLM's abilities in FOL reasoning by fully leveraging LLMs' specialized mathematical knowledge or utilizing the formal compiler effectively. Our work addresses this gap through axiom-driven strategy diversification and sub-proposition error feedback.

\section{Preliminary \& Motivation}

\subsection{Preliminary}
We treat proof generation as a sequence-to-sequence task. Given a formal FOL theorem ($x$), which includes relevant axioms that describe the features of the concepts mentioned in the theorem, our goal is to generate a formal proof ($y$) that can be automatically verified by the formal compiler $compile(.)$ \citep{de2015lean}. A proof is correct if it produces no error message, denoted as $compile(y)=pass$. Given a set of theorems $\{x_i\}^N_{i=1}$, the optimization goal for this task is to prove as many theorems as possible.

The objective function can be defined as Eq. \ref{objective}:
\begin{equation}
\label{objective}
max\sum_{i=1}^{N} \mathbb{I}(compile(y_i) = pass),
\end{equation}

where $N$ is the total number of theorems attempted, $y_i$ is the $r$-th proof for theorem $x_i$, and $\mathbb{I}$ is an indicator function that equals 1 if the proof is correct and 0 otherwise.

\subsection{Motivation}

\begin{figure}[htbp]
\centering
  \includegraphics[width=\columnwidth]{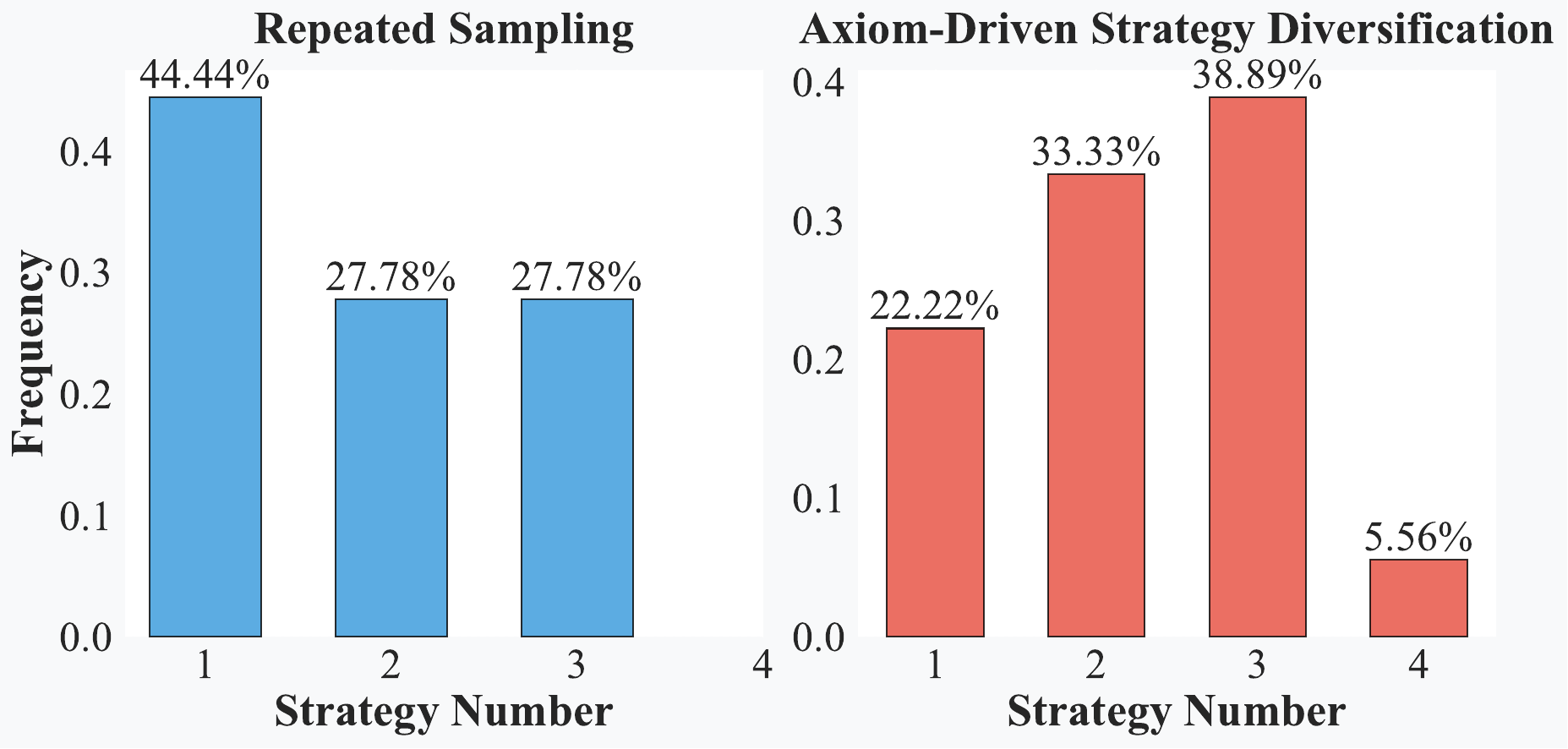}
  \caption{Comparison of strategy number distribution for six generated solutions tested on Claude 3.5.}
  \label{motivation}
\end{figure}

\textbf{Strategy Diversity}: We first explore the effect of strategy diversity on LLM's ability for FOL theorem proving tasks. Figure~\ref{motivation} reveals that repeated sampling often yields repetitive proof strategies. Since FOL deduction relies on the stepwise application of logical rules and relevant axioms or lemmas, the lack of strategy diversity will severely restrict the search space for LLMs to discover valid solutions. To mitigate this homogeneity, we experimentally investigated whether explicitly guiding LLMs to prioritize distinct axiom combinations during proof generation could break this uniformity. Our experiments demonstrate that such targeted axiom-focused prompting significantly diversifies the generated strategies under fixed computational budgets (Figure~\ref{motivation}), unlocking latent reasoning pathways. This finding motivates our proposed axiom-driven strategy diversification module, which systematically exploits axiom relevance to enhance exploration while maintaining logical coherence. Examples of diverse strategies generated by focusing on different sets of axioms are shown in Appendix~\ref{app:axiom-driven-example}.

\begin{figure}[t]
\centering
  \includegraphics[width=\columnwidth]{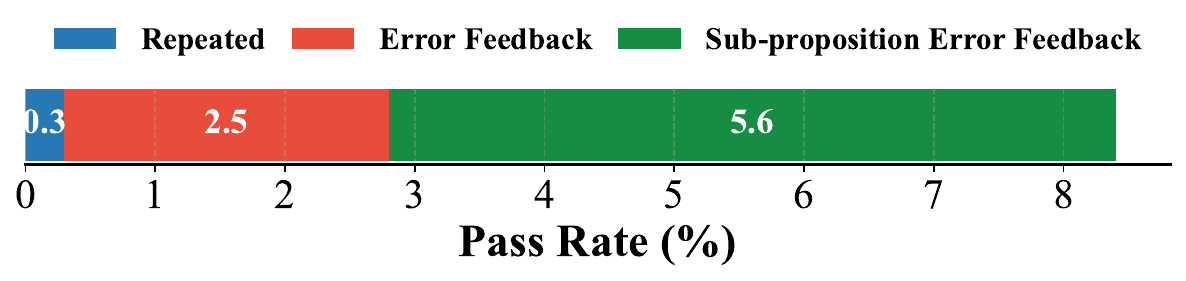}
  \caption{Pass rate on FOL theorem proving tasks using repeated sampling, error feedback, and sub-proposition error feedback. The longer rectangle is preferred.}
  \label{motivation-2}
\end{figure}

\begin{figure*}[t]
\centering
  \includegraphics[width=\textwidth]{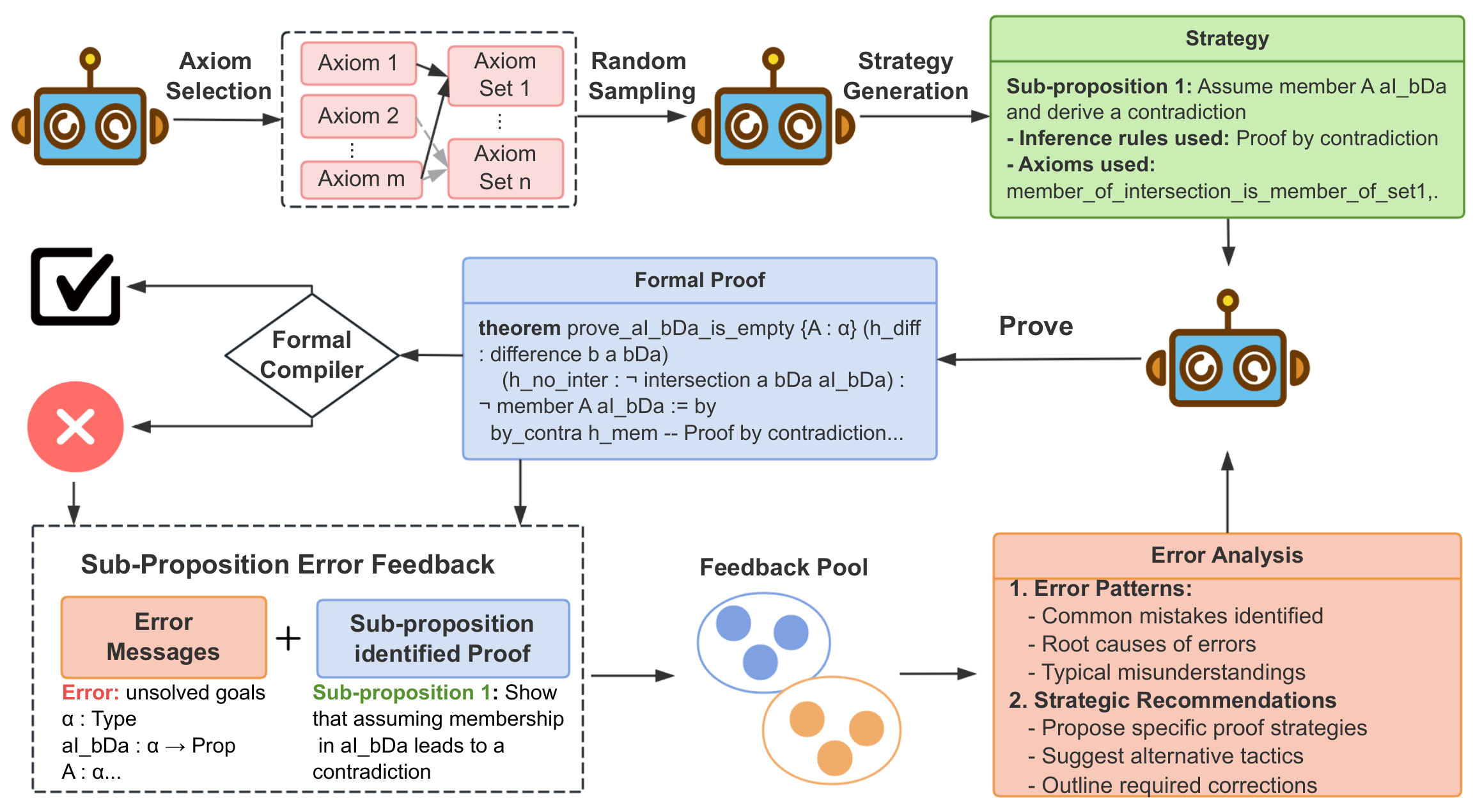}
  \caption{The overall pipeline of our proposed method. Given a conjecture, our method first applies axiom-driven strategy diversification to construct an axiom tree. Then, an axiom set is sampled from the second-level axiom tree for strategy generation. A proof is then generated based on this strategy. Incorrect proofs are labeled with sub-propositions and error messages from the formal compiler and placed into a feedback pool. Finally, an analysis of error patterns is conducted to provide strategic recommendations for iteratively refining the next round of proving.}
  \label{architecture}
\end{figure*}

\textbf{Cascading Error:} Another key factor influencing the reasonableness of FOL proofs is the cascading error (see examples in Appendix~\ref{app:cascading-error}). That's because simply providing LLMs with error messages from the formal compiler yields minimal improvement since resolving such errors demands revisions to the entire proof. To address this issue, we explore the effectiveness of mapping errors to specific sub-propositions within the proof, as shown in Figure~\ref{motivation-2}. From the results, we can observe that the sub-proposition error feedback demonstrated significant enhancement over direct error feedback, motivating our proposal for a sub-proposition error feedback module to target corrections and mitigate cascading failures by linking errors to their corresponding logical components.

\section{Method}

\subsection{Overall Framework}
In this section, we elaborate on the FOL theorem-proving framework that comprises two key components: axiom-driven strategy diversification and sub-proposition error feedback. \textbf{(i)} The axiom-driven strategy diversification aims to encourage the LLM to explore different ways of proving the theorem. To achieve this goal, given a theorem $x$, we can construct a \textit{k}-wise combinatorial axiom tree to update strategies, which are executed through two or three times of revisions to prevent the LLM from getting stuck in the same incorrect reasoning; \textbf{(ii)} The sub-proposition error feedback aims to further ensure the reasonability of reasoning chains during theorem proving, which takes advantage of back-propagating error messages from previous failed proofs. We create sub-proposition error feedback that enhances self-correction by linking these error messages to sub-propositions. The model learns from sub-proposition errors of earlier attempts at each revision time. The overall framework is illustrated in Figure~\ref{architecture}.

\subsection{Axiom-Driven Strategy Diversification}

To address the adaptive strategy starvation, we aim to expand the strategy search space by constructing a \textit{k}-wise combinatorial axiom tree. Similar techniques can also be shown in \citet{wang2024planning}, which focuses on the LLM planning. This tree allows LLMs to systematically explore various strategies, improving their success rate.

Denote the LLM as $\theta$ and $p_\theta$ as the probability distribution from the LLM. We can initially generate a set of first-level axioms based on the context and the conjecture. The $O = \{o_1, o_2, \ldots, o_M\}$ are defined as axioms, where $O$ is sampled from the distribution $p_\theta(\cdot \mid x)$, $o_i$ denotes an individual axiom, and $M$ is the number of axioms. 

Specifically, we employ the \textit{k}-wise combinatorial generation tree, where the second-level axioms $O'$ are generated based on these first-level axioms. Each second-level axiom $o'_s$ is a leaf node of the \textit{k}-wise combinatorial generation tree and can be derived from one possible \textit{k}-wise combination of the first-level axioms. We can denote the second-level axioms as Eq. \ref{second_level}, where $S_k$ stands for the indexes of all possible \textit{k}-wise combinations from $M$ axioms in the first-level as Eq. \ref{sk}. The number of elements in $\mathcal{S}_k$, $\binom{M}{k}$, represents the number of ways to choose $k$ elements from a set of $M$ distinct elements. Strategy $\mathcal{P}$ is generated using a new second-level axiom set $o'_{s}$. We can donate the strategy as Eq.~\ref{strategy}.

\resizebox{0.8\linewidth}{!}{
\begin{minipage}{\linewidth} 
\begin{align}
    \label{second_level}
    \mathcal{O}' &= \Bigl\{ o'_{s} \,\Big|\, o'_{s} \sim p_\theta\bigl(\cdot \mid \{o_{s_i}\}^k_{i=1};\, x \bigr),\ s \in \mathcal{S}_k \Bigr\} \\
    \label{sk}
    \mathcal{S}_k &= \Bigl\{ s = (s_1, \dots, s_k) \,\Big|\, 1 \leq s_1 < \cdots < s_k \leq M \Bigr\} \\
    \label{strategy}
    \mathcal{P} &\sim p_\theta(\ \cdot \ \big| x;\ o'_{s}) 
\end{align}
\end{minipage}
}

\subsection{Sub-proposition Error Feedback}
To address the cascading error propagation inherent in formal proof correction, we leverage error feedback from formal verification compilers to iteratively refine LLM-generated proofs. However, FOL theorem proofs have numerous sub-propositions linked using logical connections like conjunction $\wedge$ and disjunction $\vee$. Directly applying word-level error messages generated by the formal compiler may not lead LLMs to create the linkage modifications between sub-propositions of first-order logic, seriously damaging the thorough proof correction. Thus, we propose the sub-proposition-level error feedback where the error messages are strictly aligned with the sub-propositions of the proof.  

Denote the set of all previous $r-1$ failed attempts as $E=\{E_1, E_2,\dots,E_{r-1}\}$. Each attempt contains a formal proof of the statement $y_i$ and corresponding error messages $e_i=Compiler(y_i)$, where $Compiler(\cdot)$ is the formal compiler. We represent the aligned proofs $y'$ using inline comments, placing sub-proposition annotations before the code block and error messages after the corresponding error line. $y'_i$ is generated by an sub-proposition annotator $L$, $y'_i = L(E_i)$. An analyzer $A$ examines mistakes at the sub-propositional level, offering insights for the $r$-th revision $I_r$ into error patterns and suggesting strategies for improvement. We denote $I_r$ as Eq. \ref{I_r}, where $r$ stands for the current revision time.

\begin{equation}
\label{I_r}
I_r = A(\  x; \ \{y'_i\}^{r-1}_{i=1} \ ) 
\end{equation}

\noindent The proof of current revision time $y_r$ is generated as Eq. \ref{current_revision}:
\begin{equation}
\label{current_revision}
y_r \sim p_\theta(\ \cdot \ \big| x;\ I_r; \{E_j\} ^{r-1}_{j=1}) 
\end{equation}

where $p_\theta$ represents the generative model, $x$ denotes the theorem, $I_r$ signifies the insight, and $\{E_j\} ^{r-1}_{j=1} $ represents the collection of previous proofs with corresponding error messages generated by the compiler.

\section{Experiment}

\subsection{Experimental Setup}

\textbf{Baselines:} We adopt two well-known inference-stage solutions as comparisons to display the effectiveness of our method: (i) Repeated sampling (Repeated), where the LLM generates a correct proof for a theorem until it either reaches the maximum attempts or passes the formal compiler; (ii) Subgoal-based demonstration learning (Subgoal), which breaks down the theorem into subgoals in natural language and selects relevant examples for in-text demonstration learning \citep{pmlr-v235-zhao24h}. \\

\noindent \textbf{Evaluation Metric:} Following \citet{pmlr-v235-zhao24h}, we select the cumulative pass rate as the metric for evaluation, which is the proportion of theorems solved at least once. A large pass rate is preferred. \\

\begin{figure}[htbp]
\centering
  \includegraphics[width=\columnwidth]{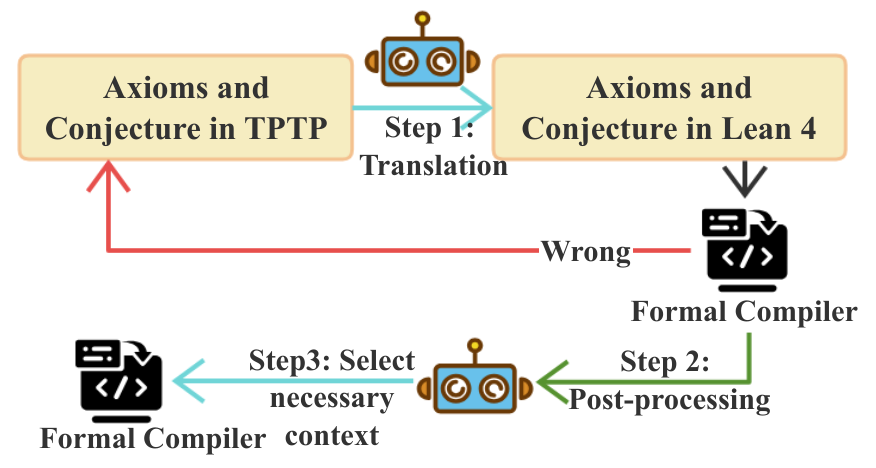}
  \caption{TPTP revision pipeline. }
  \label{dataset-construction}
\end{figure}

\noindent \textbf{Evaluation Dataset:} Due to the lack of a mathematical evaluation benchmark with multi-step FOL deductions, we construct it as follows: \textbf{(i) TPTP Revised Dataset.}  We converted 324 FOL problems from the TPTP format ~\citep{sutcliffe2017tptp} to the Lean 4 format to support LLM proving. Specifically, we utilize LLMs to translate axioms and conjectures from TPTP to Lean 4 format, leveraging their exceptional expertise in Lean 4. Similar to prior autoformalization approaches~\citep{zhang-etal-2024-consistent, yang-etal-2024-harnessing}, the dataset construction pipeline is illustrated in Figure~\ref{dataset-construction}. It involves three key steps: \textit{Step 1: Lean 4 Format Translation:} We employ DeepSeek-V3 to convert conjectures and their associated axioms from TPTP to Lean 4 format. The translation prompt is detailed in Appendix~\ref{prompt-tptp2lean}, and each translated example is verified with the Lean 4 compiler~\citep{de2015lean}. \textit{Step 2: Post-processing:} To facilitate LLM proving, we separate conjecture definitions from their context, add necessary import statements, and manually review the content for quality assurance. \textit{Step 3: Context Optimization:} To improve LLM comprehension, we use DeepSeek-V3 to retain only essential contextual elements. The formal compiler then verifies the simplified problems; \textbf{(ii) Manually Collected Dataset:}  We curated a new dataset featuring 123 problems to cover various topics in the FOL theorem proving theme. Specifically, we manually collect theorems from academic papers and discrete mathematics textbooks. These theorems were converted to LaTeX and verified as valid first-order logic statements. They were then transformed from LaTeX to Lean 4 format using DeepSeek-V3, with up to 60 attempts. The dataset emphasizes intuitionistic logic, set theory, and computability, covering realizability, model theory, substitution, tautologies, and relationships between logical systems. Two human verifiers reviewed the annotations and corrected any inaccuracies.\\

\begin{table}[!th]
  \centering
  \small
  \fontsize{8.5pt}{8.5pt}\selectfont 
    \setlength{\tabcolsep}{3.4pt}
    \renewcommand{\arraystretch}{1}
    \resizebox{\columnwidth}{!}{
   \begin{tabular}{c|ccccc}
   
    \toprule
    \textbf{Dataset} & \textbf{Creation} & \textbf{Domain}  & \textbf{Formal} & \textbf{Division}\\
    \midrule

    RuleTaker~\citep{clark2020transformers} & Synthetic & Real-world &\ding{55} &\ding{55} \\
    ProofWriter~\citep{tafjord-etal-2021-proofwriter} & Synthetic & Real-world & \ding{55} & \ding{55} \\
    LogicNLI~\citep{tian-etal-2021-diagnosing} & Synthetic    & Real-world & \ding{55} & \ding{55} \\
    ProntonQA~\citep{saparov2023language} & Synthetic   & Real-world & \ding{51}   & \ding{55} \\
    FOLIO~\citep{han2024folionaturallanguagereasoning} & Manual & Real-world  &\ding{51} &\ding{55}  \\
    ProverQA~\citep{qi2025largelanguagemodelsmeet} & Synthetic   & Real-world & \ding{51}   & \ding{55} \\

    \midrule
     
    \textbf{Our Proposed Dataset}   & Synthetic \& Manual  & Mathematics & \ding{51} & \ding{51}\\
    \bottomrule
\end{tabular}
}

\caption{Comparison between our mathematical FOL reasoning dataset and existing FOL datasets. "Formal" indicates the inclusion of a formal format, while "Division" refers to the subcategories within the dataset. }
\label{comp-datasets}
\end{table}

\noindent \textbf{Implementation Details:} We employ Claude 3.5 Sonnet~\citep{anthropic2024claude35} and DeepSeek-Prover-V2-7B~\citep{openai2024o1card} as the LLMs. For FOL theorem proving tasks, we utilize a 2-level, 2-wise combinatorial axiom tree, generating three to five axiom nodes at the first level. The maximum number of attempts is set to 10. Axiom-driven diversification is applied in the 4th and 7th revisions. 

\subsection{Performance Comparison}

\begin{table*}[htbp]
  \centering
  \small
  \fontsize{8.5pt}{8.5pt}\selectfont 
    \setlength{\tabcolsep}{3.4pt}
    \renewcommand{\arraystretch}{0.5}
    \resizebox{\textwidth}{!}{
    \begin{tabular}{c|c|ccccccccc|c}
    \toprule

\textbf{Models} & \textbf{Methods} & \textbf{FLD1} & \textbf{FLD2} & \textbf{GEO6} & \textbf{GEO8} & \textbf{GEO9} & \textbf{GRP5} & \textbf{NUM9} & \textbf{KRS1} & \textbf{SET1} & \textbf{Avg.}  \\
\midrule

\multirow{3}{*}{Claude 3.5} & Repeated~\citep{brown2024largelanguagemonkeysscaling} & 0.0\% & 0.0\% & 0.0\% & 0.0\% & 0.0\% & 0.0\% & 0.0\% & 1.5\% & 0.0\% & 0.2\% \\
& Subgoal~\citep{pmlr-v235-zhao24h} & 5.2\% & 3.1\% & 4.5\% & \textbf{4.9\%} & 12.5\% & 0.0\% & 0.0\% & 3.0\% & 0.0\% & 3.7\% \\

& \textbf{DREAM(Ours)} & \textbf{14.3\%} & \textbf{12.5\%} & \textbf{13.6\%} & 0.0\% & 0.0\% & \textbf{20.0\%} & \textbf{5.6\%} & \textbf{3.0\%} & \textbf{22.2\%} & \textbf{10.1\%} \\

\midrule

\multirow{3}{*}{DeepSeek-Prover-V2-7B} & Repeated~\citep{brown2024largelanguagemonkeysscaling} & 1.3\% & 0.0\% & 6.8\% & 7.3\% & 0.0\% & 10.0\% & 0.0\% & \textbf{1.5\%} & 11.1\% & 4.2\% \\
& Subgoal~\citep{pmlr-v235-zhao24h} & 0.0\% & \textbf{3.1\%} & 9.1\% & 4.9\% & \textbf{25.0\%} & 20.0\% & \textbf{5.6\%} & \textbf{1.5\%} & 0.0\% & 7.7\% \\
& \textbf{DREAM(Ours)} & \textbf{3.9\%} & 0.0\% & \textbf{11.4\%} & \textbf{9.8\%} & 12.5\% & \textbf{30.0\%} & \textbf{5.6\%} & \textbf{1.5\%} & 0.0\% & \textbf{8.3\%} \\

\bottomrule

\end{tabular}
}

\caption{Performance comparison on the TPTP revised dataset.  "Avg." refers to the average pass rate (\%).}
\label{comp-method-base}
\end{table*}

\begin{table}[htbp]
  \centering
  \small
  \fontsize{8.5pt}{8.5pt}\selectfont 
    \setlength{\tabcolsep}{3.4pt}
    \renewcommand{\arraystretch}{0.5}
    \resizebox{\columnwidth}{!}{
   \begin{tabular}{c|c|c}
    \toprule
  \textbf{Models} & \textbf{Methods} & \textbf{Avg.}\\
    \midrule
    
 \multirow{3}{*}{Claude 3.5} & Repeated~\citep{brown2024largelanguagemonkeysscaling} & 32.5\%\\
    
& Subgoal~\citep{pmlr-v235-zhao24h} & 27.6\%  \\
& \textbf{DREAM(Ours)} & \textbf{41.5\%}\\

    \midrule
    
\multirow{3}{*}{DeepSeek-Prover-V2-7B} & Repeated~\citep{brown2024largelanguagemonkeysscaling} & 12.2\%\\
& Subgoal~\citep{pmlr-v235-zhao24h} & \textbf{22.8\%}\\
& \textbf{DREAM(Ours)} & 21.1\% \\

    \bottomrule
\end{tabular}
}

\caption{Performance comparison on the manually collected dataset. "Avg." denotes the average pass rate (\%).}
\label{results-manual}
\end{table}

As shown in Table~\ref{comp-method-base} and Table~\ref{results-manual}, we have the following key findings:

(i) Despite extensive training on formal mathematical proving materials, the LLMs tested in our dataset still performed poorly, highlighting the challenging nature of our proposed dataset. Claude 3.5 achieves a mere 0.2\% pass rate, while DeepSeek-Prover-V2-7B reaches only 4.2\%. The models perform relatively better on the manually collected dataset, which encompasses a broader range of mathematical topics and includes shorter contexts with fewer logical restrictions. This observation suggests that LLMs struggle with reasoning under strict logical constraints, such as FOL rules and axioms. 

(ii) Our proposed DREAM significantly outperforms other methods on the FOL theorem proving task, achieving an average pass rate of 10.1\% using Claude 3.5 and 8.3\% using DeepSeek-Prover-V2-7B. Specifically, DREAM demonstrated its superior performance across all domains, showing its efficiency in FOL theorem proving. The repeated sampling method underperforms because of its limited search space on strategies, which prevents it from exploring more possible solutions, leading to repeated failures on the same errors. The subgoal-based demonstration learning method has introduced subgoal decompositions and demonstration examples. However, this approach overlooks FOL logic's non-modular error propagation characteristic, addressing errors only within specific modules. In comparison, our method uses a \textit{k}-wise combinatorial axiom tree, allowing for systematic exploration of strategies. The specially designed axiom-driven strategy diversification has guaranteed its stable performance by systemically exploring different strategies. In contrast, the sub-proposition error feedback designed according to the feature of first-order logic stably guides the LLMs to the correct proof pathway. These mechanisms have resulted in the strong generalization ability of our method, making it well-suited for more diverse FOL theorem-proving tasks.

\begin{figure}[htbp]
    \centering
    \begin{subfigure}{0.47\columnwidth}
        \includegraphics[width=\linewidth]{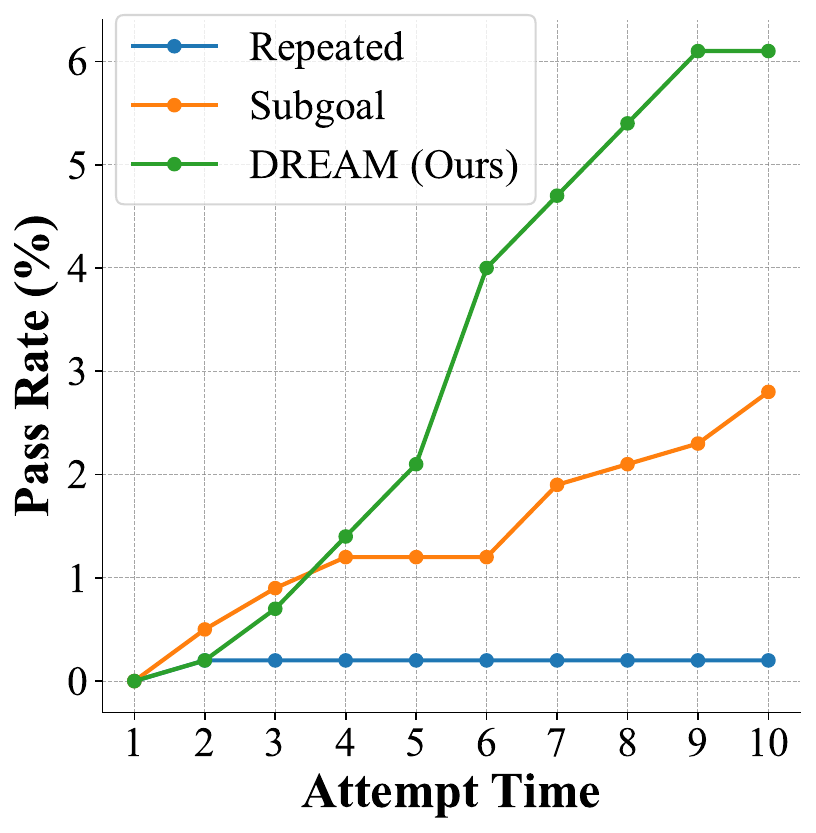}
    \end{subfigure}%
    \begin{subfigure}{0.47\columnwidth}
        \includegraphics[width=\linewidth]{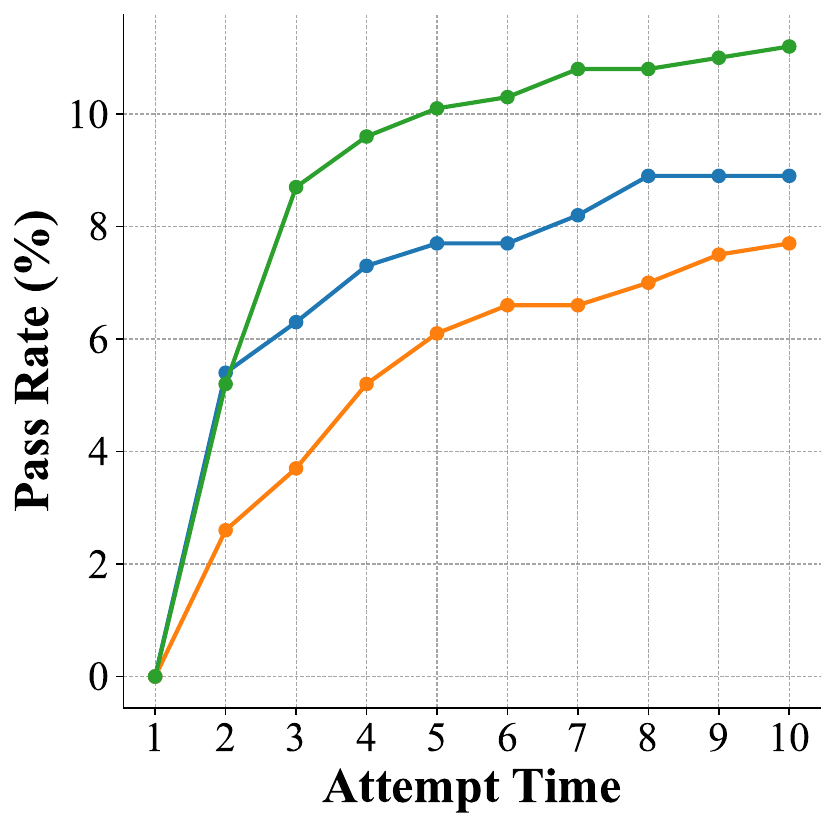}
    \end{subfigure}
    \caption{Passing rate comparisons on Claude 3.5 for various methods on the TPTP revision dataset (left) and the manually collected dataset (right) across attempts. The x-axis indicates the attempt number. Our proposed method achieves the highest passing rate starting from the fourth attempt on the TPTP revised dataset and the third attempt on the manually collected dataset.}
    \label{revision-num-dataset}
\end{figure}

(iii) We also monitor the pass rates of various approaches, as shown in Figure~\ref{revision-num-dataset}, where we adopt different methods on our dataset using Claude 3.5. Initially, our method may not have performed as well as the subgoal-based demonstration learning method on the TPTP revised dataset. However, after the fourth revision, it began to show significant improvement. This trend suggests that our approach benefits from the iterative learning process, where each revision builds on the last. Our method achieves the top rank significantly after the second revision on the manually collected dataset, while the Subgoal-based demonstration learning method ranks the lowest. This result demonstrates the strong generalization ability of our method across different types of problems.

\subsection{Ablation Studies}

\begin{table*}[!t]
  \centering
  \small
  \fontsize{8.5pt}{8.5pt}\selectfont 
    \setlength{\tabcolsep}{3.4pt}
    \renewcommand{\arraystretch}{0.5}
    \resizebox{0.9\textwidth}{!}{
    \begin{tabular}{c|cc|ccccccccc|c}
    \toprule

\textbf{Models} & \textbf{SD} & \textbf{SE} & \textbf{FLD1} & \textbf{FLD2} & \textbf{GEO6} & \textbf{GEO8} & \textbf{GEO9} & \textbf{GRP5} & \textbf{NUM9} & \textbf{KRS1} & \textbf{SET1} & \textbf{Avg.}  \\
 \midrule

 \multirow{4}{*}{Claude 3.5} & - & - & 0.0\% & 0.0\% & 0.0\% & 0.0\% & 0.0\% & 0.0\% & 0.0\% & 1.5\% & 0.0\% & 0.2\%  \\

 & \checkmark & - & 13.0\%   &0.0\%  &9.1\%  &2.4\%   &0.0\%  &0.0\% &0.0\% &3.0\% &11.1\% &5.6\% \\
 & - & \checkmark & 3.9\% & 3.1\% & 11.4\% & 2.4\% & \textbf{12.5\%} & \textbf{20.0\%} & 0.0\% & 3.0\% & \textbf{33.3\%} & 9.9\%  \\
 
 & \checkmark & \checkmark & \textbf{14.3\%} & \textbf{12.5\%} & \textbf{13.6\%} & 0.0\% & 0.0\% & \textbf{20.0\%} & \textbf{5.6\%} & \textbf{3.0\%} & 22.2\% & \textbf{10.1\%} \\

\midrule

 \multirow{4}{*}{DeepSeek-Prover-V2-7B} & - & - & 1.3\% & 0.0\% & 6.8\% & 7.3\% & 0.0\% & 10.0\% & 0.0\% & \textbf{1.5\%} & \textbf{11.1\%} & 4.2\% \\

 & \checkmark & - & 0.0\%   &0.0\%  &2.3\%  &4.9\%   &\textbf{12.5\%} &10.0\% &2.8\% &1.5\% &0.0\% &3.8\% \\
 & - & \checkmark & 0.0\% & 0.0\% & 2.3\% & 2.4\% & 0.0\% & \textbf{20.0\%} & 2.8\% & 1.5\% & \textbf{0.0\%} & 3.2\%  \\
 
 & \checkmark & \checkmark & \textbf{3.9\%} & 0.0\% & \textbf{11.4\%} & \textbf{9.8\%} & \textbf{12.5\%} & \textbf{30.0\%} & \textbf{5.6\%} & \textbf{1.5\%} & 0.0\% & \textbf{8.3\%} \\

\bottomrule

\end{tabular}
}

\caption{Ablation results on TPTP revised dataset. "SD" stands for axiom-driven strategy diversification. "SE" stands for sub-proposition error feedback. "Avg." refers to the average pass rate (\%).}
\label{ablation}
\end{table*}

To further understand the effectiveness of our proposed DREAM, we analyze the factors that influence its efficiency based on the TPTP revised dataset. Overall, as shown in Table~\ref{ablation}, we can observe that DREAM achieves the highest performance across most domains. This performance underlines the effectiveness of various modules as follows. 

\textbf{Analysis on axiom-driven strategy diversification.} The effectiveness of the axiom-driven strategy diversification is evident when we analyze its removal from our full method. Without it, the pass rate decreased from 10.1\% to 9.9\% for Claude 3.5 and from 8.3\% to 3.2\% for DeepSeek-Prover-V2-7B. Except for domains like GEO8, GEO9, and SET1, its absence generally leads to lower pass rates. However, using this mechanism alone does not guarantee improved performance, as Claude 3.5's pass rate increased, while DeepSeek-Prover-V2-7B's decreased.

\textbf{Analysis on Sub-proposition Error Feedback.} The lack of sub-proposition error feedback has resulted in a significant decrease in the average pass rate, dropping from 10.1\% to 5.6\%, with all domains showing a notable decline. This decline may be because the sub-propositions in the proofs allow LLMs to analyze the explored strategies, refine the decomposition of the main theorem's sub-propositions, and offer targeted revision insights.

\subsection{Discussion about Background Restrictions} 

In our experiments, we also observed an interesting phenomenon related to the background restrictions for the axiom. Specifically, we can remove standard mathematical axioms (FLD, GEO, GRP, NUM, SET) to analyze LLMs' internal abilities for knowledge recall. Using 32 axiom-free problems (4 per domain from 8 TPTP domains), Claude 3.5's success rate increased while DeepSeek-Prover-V2-7B's declined compared to axiom-dependent scenarios (Table~\ref{results-non-axiom}). This contrast suggests Claude owns broader mathematical knowledge as a general LLM, flexibly applying familiar axioms. However, the training of DeepSeek-Prover-V2-7B relies on complete proofs related to predefined backgrounds, leading to a model excelling in structured contexts but struggling when axioms are removed. The divergence highlights how training data (specialized proofs vs general knowledge) shapes FOL problem-solving approaches.

\subsection{Case Studies} 
\begin{figure}[h]
\centering
  \includegraphics[width=\columnwidth]{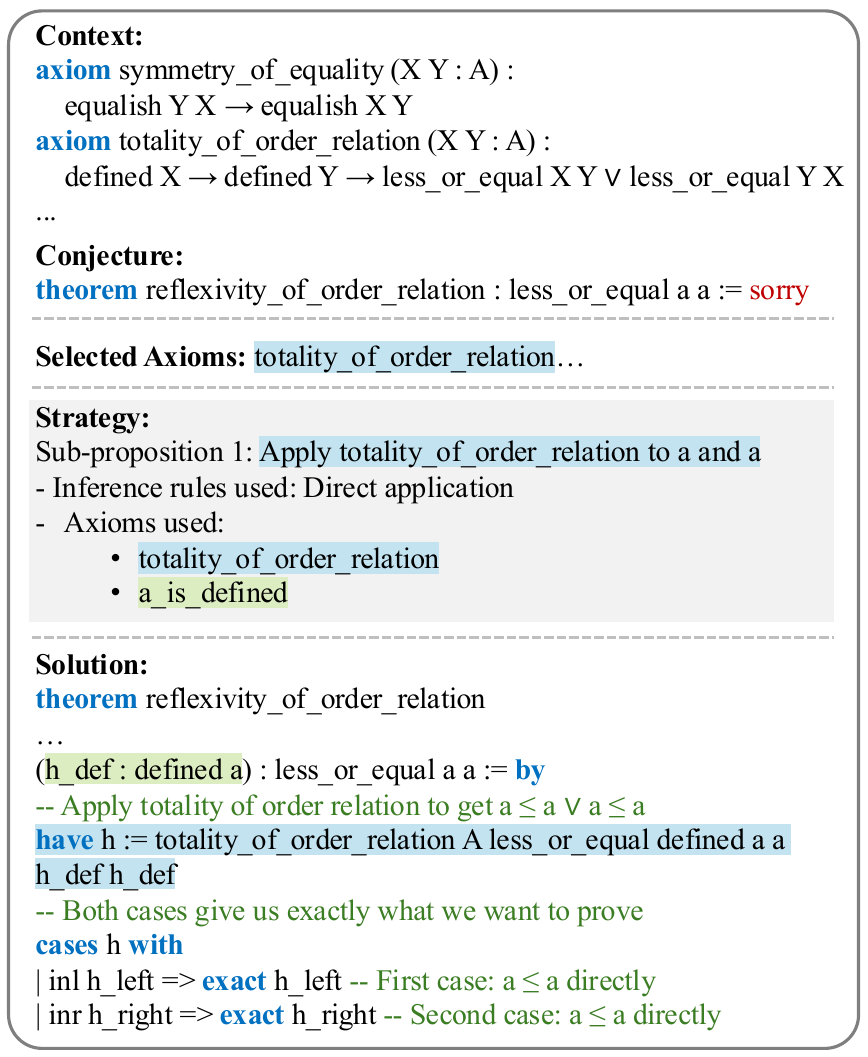}
  \caption{Case study: The effect of strategy diversification on proving a theorem for Claude 3.5.}
  \label{case-study}
\end{figure}

\begin{figure}[h]
\centering
  \includegraphics[width=\columnwidth]{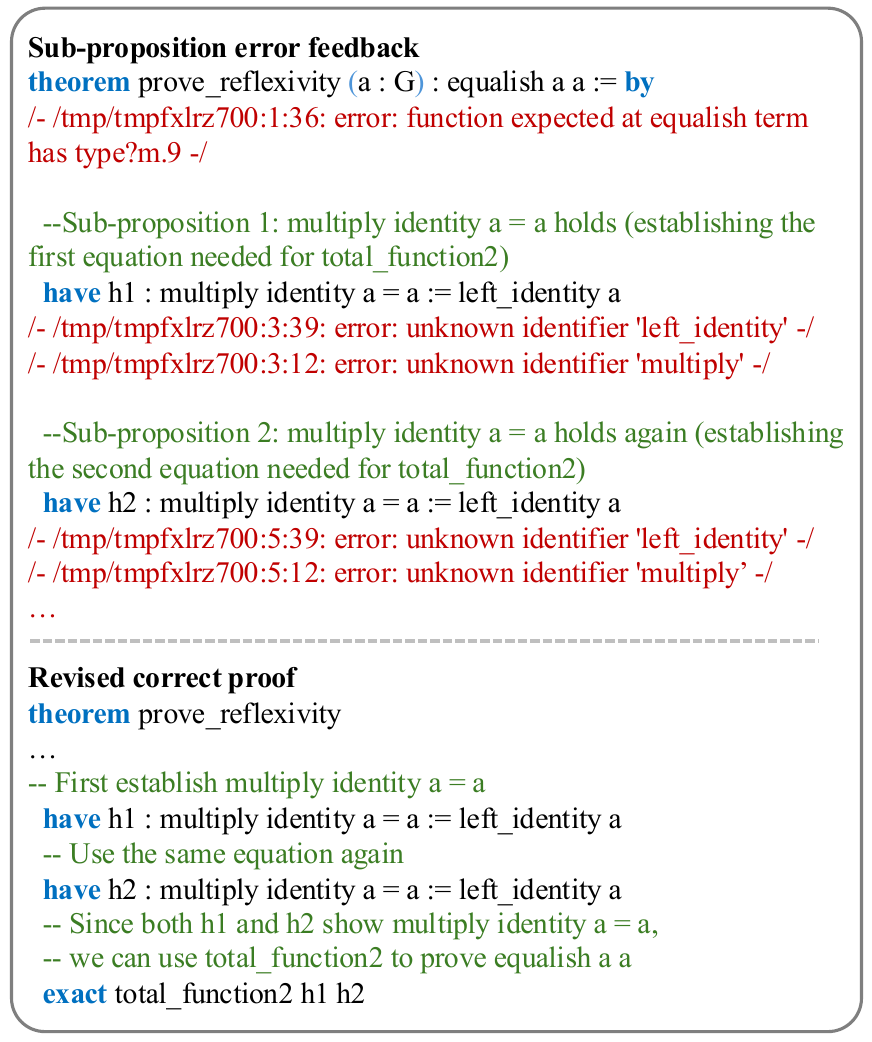}
  \caption{Case study: The effect of sub-proposition error feedback on  proving a theorem for Claude 3.5.}
  \label{case-study-error}
\end{figure}

We provide two cases related to strategy diversity and sub-proposition error feedback to visualize the effectiveness of our method in solving FOL theorem proving problems. Figure~\ref{case-study} shows that our method successfully generates the correct proof after applying the axiom-driven strategy diversification. The strategy is derived from the totality\_of\_order\_relation axiom's key feature of providing a disjunctive conclusion. This strategy, "Apply totality\_of\_order\_relation to a and a", is directly implemented in the solution through have h := totality\_of\_order\_relation A less\_or\_equal defined a a h\_def h\_def, where both resulting cases yield the desired reflexivity property. Figure~\ref{case-study-error} illustrates how incorporating sub-proposition error feedback enables the LLMs to make high-level modifications, resulting in successful proofs.

\begin{table}[!th]
  \centering
  \small
  \fontsize{8.5pt}{8.5pt}\selectfont 
    \setlength{\tabcolsep}{3.4pt}
    \renewcommand{\arraystretch}{0.5}
    \resizebox{\columnwidth}{!}{
   \begin{tabular}{c|c|cc}
    \toprule
    \textbf{Models} &\textbf{Methods} & \textbf{w/o Background} & \textbf{w Background}\\
    \midrule

    \multirow{3}{*}{Claude 3.5}  & Repeated~\citep{brown2024largelanguagemonkeysscaling} & 12.5\%  & 0.0\%\\
    
    & Subgoal~\citep{pmlr-v235-zhao24h} & 18.8\% & 3.1\% \\
    
    & \textbf{DREAM(Ours)} & 21.9\% & 12.5\% \\

    \midrule
    \multirow{3}{*}{DeepSeek-Prover-V2-7B} & Repeated~\citep{brown2024largelanguagemonkeysscaling} & 0.0\% & 9.4\%\\
    & Subgoal~\citep{pmlr-v235-zhao24h} & 3.1\% & 6.3\%\\
    & \textbf{DREAM(Ours)} & 3.1\% & 9.4\% \\

    \bottomrule
\end{tabular}
}

\caption{The exploration is related to the background constraint for axioms. We randomly selected 32 samples from the TPTP revised FOL-based dataset.}
\label{results-non-axiom}
\end{table}

\section{Conclusion}
 
This work first advances LLMs' mathematical reasoning abilities via first-order theorem proving. Through detailed TPTP revision and manual collection, we curated a challenging FOL theorem-proving dataset that uncovers the drawbacks of existing formal LLMs (e.g., DeepSeek-Prover-V2-7B). Moreover, we propose the DREAM framework, a novel inference stage solution incorporating axiom-driven strategy diversification and sub-proposition error feedback for efficient FOL theorem proving. Our approach effectively addresses the limitations of LLMs in handling mathematical first-order logic-proving tasks in formal formats. Extensive experiments can verify the effectiveness of DREAM over previous methods for this challenging and complex reasoning task.

\section{Limitations}
Our experiments demonstrate the effectiveness of DREAM in enhancing LLMs' performance in FOL-based theorem-proving tasks, more diverse FOL-based mathematical tasks could be considered in the future. Additionally, the experimental results show a consistent increase in performance, even by the 10th revision. However, due to resource limitations, we have no chance to extend the experiment to identify our method's saturation point. Future research should also account for the model’s internal structured reasoning patterns~\citep{wen2025thinkpatterns}. In addition to performance, the ethical and societal acceptability, such as safety, honesty and value, should also be incorporated to enhance the controllability and reliability of reasoning ~\citep{cao2025safelawbench,yang2025mix, ju-etal-2025-benchmarking}. 


\section*{Acknowledgments}
This work is funded in part by the HKUST Start-up Fund (R9911), Theme-based Research Scheme grant (T45-205/21-N), the InnoHK initiative of the Innovation and Technology Commission of the Hong Kong Special Administrative Region Government, and the research funding under HKUST-DXM AI for Finance Joint Laboratory (DXM25EG01).

\bibliography{custom}

\clearpage
\appendix

\label{sec:appendix}

\section{Pseudocode}

The pseudocode for our proposed method is presented in Table~\ref{pseudocode}.

\begin{table}[htbp]

\resizebox{\columnwidth}{!}{
\begin{tabular}{l}
\toprule
\textbf{Algorithm: ProveTheorem} \\
\midrule
\textbf{Input}: Conjecture $x$, LLM $\theta$, max revision $R$ \\
\textbf{Output}: Proof $y$ \\
\midrule
$O \leftarrow$ GenerateFirstLevelAxioms($x$, $\theta$) \\
$O' \leftarrow$ GenerateSecondLevelAxioms($O$, $x$, $\theta$, $k$) \\
$E \leftarrow \{\}$ // Initialize error collection \\
\textbf{for} $r \leftarrow 1$ \textbf{to} $R$ \textbf{do} \\
\quad $o'_s \leftarrow$ SelectAxioms($O'$) \\
\quad \textbf{if} $r = 0$ \textbf{then} \\

\quad\quad $y_r \leftarrow$ GenerateInitialProof($x$, $\theta$) \\

\quad \textbf{if} $r = 4$ or $r = 7$ \textbf{then} \\
\quad\quad $o' \leftarrow$ SelectSecondLevelAxioms($O'$) \\
\quad\quad $P \leftarrow$ GenerateStrategy($x$, $o'$) \\
\quad\quad $y_r \leftarrow$ GenerateProofBasedOnStrategy($x$, $P$, $\theta$) \\

\quad \textbf{else} \\
\quad\quad $I_r \leftarrow$ AnalyzeWithFeedback($x$, $o'_s$, $\{y'_i\}^{r-1}_{i=1}$) \\
\quad\quad $y_r \leftarrow$ GenerateRevisedProof($x$, $I_r$, $E$, $\theta$) \\
\quad \textbf{if} compile($y_r$) = pass \textbf{then} \\
\quad\quad \textbf{return} $y_r$ \\
\quad $e_r \leftarrow$ compile($y_r$) \\
\quad $y'_r \leftarrow$ AnnotateProof($y_r$, $e_r$) \\
\quad $E \leftarrow E \cup \{(y_r, e_r)\}$ \\
\textbf{return} NULL \\
\bottomrule
\end{tabular}
}
\caption{Core pseudocode of DREAM for FOL theorem proving.}
\label{pseudocode}
\end{table}

\section{Dataset Statistics}

The domain distribution from the TPTP library is shown in Table~\ref{tptp-data}
\begin{table}[h]
  \centering
  \fontsize{8.5pt}{8.5pt}\selectfont 
    \setlength{\tabcolsep}{3.4pt}
    \renewcommand{\arraystretch}{0.9}
    \resizebox{\columnwidth}{!}{
  \begin{tabular}{l|lcc}
    \toprule
    \textbf{Domain}  & \textbf{Description} & \textbf{TPTP}     &\textbf{Lean4}\\
    \midrule
    FLD1    & Field Theory           & 136    & 77   \\
    FLD2    & Field Theory    & 143     & 32 \\
    GEO6    & Geometry    & 97     & 44 \\
    GEO8    & Geometry   & 57     & 41 \\
    GEO9    & Geometry   & 43     & 8 \\
    GRP5    & Group Theory   & 10    & 10 \\
    NUM9    & Number Theory     & 36    & 36 \\
    KRS1    & Knowledge Representation  & 94    & 67 \\
    SET1    & Set Theory     & 11 & 9 \\
    \midrule
    Total   & & 627 & 324 \\

    \bottomrule
  \end{tabular}
  }
  \caption{\label{tptp-data}
    First-order theorems extracted from the TPTP library \citep{sutcliffe2017tptp}. 
  }
\end{table}

\section{Data Quality Control}
\label{sec:human-agreement}

Similar to prior autoformalization approaches like DeepSeek-Prover-V2~\citep{ren2025deepseekproverv2advancingformalmathematical, zhang2024consistent, lu2024process}, we employ Lean verification to ensure logical accuracy and mitigate potential biases in LLM-assisted dataset creation.

To assess the reliability of the manual verification stage, we conducted a human annotation study measuring inter-annotator agreement. Two experts in formal reasoning independently evaluated 40 problems—20 from the manually collected corpus and 20 from TPTP-Revised—for correctness and logical consistency of theorem statements and their Lean 4 formalizations. As shown in Table~\ref{tab:human-agreement}, the overall problem-wise agreement was 82.5\%. This is consistent with the 81\% human agreement reported by~\citet{zheng2023judging}, supporting the reliability of our dataset.

\begin{table*}[htbp]
\centering
\small
\setlength{\tabcolsep}{8pt}
\renewcommand{\arraystretch}{1.1}
\resizebox{\textwidth}{!}{
\begin{tabular}{lrrr}
\toprule
\textbf{Annotators} & \textbf{Correct Rate (Manual)} & \textbf{Correct Rate (TPTP-Revised)} & \textbf{Avg. Correct Rate} \\
\midrule
Annotator 1 & 90\% & 85\% & 87.5\% \\
Annotator 2 & 90\% & 95\% & 92.5\% \\
\midrule
Problem-wise Agreement & 85\% & 80\% & 82.5\% \\
\bottomrule
\end{tabular}
}
\caption{Problem-wise agreement between two annotators on manually collected and TPTP-revised problems.}
\label{tab:human-agreement}
\end{table*}

\section{Detailed Statistics of Dataset Construction}

We provide detailed statistics for both steps of the dataset construction pipeline in Table~\ref{tab:tptp-revised-success-rate}.

\begin{table*}[htbp]
\centering
\small
\fontsize{8.5pt}{8.5pt}\selectfont 
\setlength{\tabcolsep}{3.4pt}
\renewcommand{\arraystretch}{1.2} 

\resizebox{\textwidth}{!}{
\begin{tabular}{lcccccc}
\toprule
\textbf{Category} 
& \textbf{Original} 
& \makecell{\textbf{Success} \\ \textbf{Translations}} 
& \makecell{\textbf{Success Translation} \\ \textbf{Rate (\%)}} 
& \makecell{\textbf{Avg. Success} \\ \textbf{Translation Time}} 
& \makecell{\textbf{Success} \\ \textbf{Optimizations}} 
& \makecell{\textbf{Success Optimization} \\ \textbf{Rate (\%)}} \\
\midrule
FLD001   & 136 & 96  & 70.6 & 4.0 & 77  & 80.2 \\
FLD002   & 143 & 38  & 26.6 & 5.0 & 32  & 84.2 \\
GEO006   & 97  & 57  & 58.8 & 8.0 & 44  & 77.2 \\
GEO008   & 57  & 45  & 78.9 & 2.0 & 41  & 91.1 \\
GEO009   & 43  & 13  & 30.2 & 5.0 & 8   & 61.5 \\
GRP005   & 10  & 10  & 100.0 & 2.0 & 10  & 100.0 \\
NUM009   & 36  & 36  & 100.0 & 1.0 & 36  & 100.0 \\
KRS001   & 94  & 75  & 79.8 & 2.0 & 67  & 89.3 \\
SET001   & 11  & 9   & 81.8 & 5.0 & 9   & 100.0 \\
\midrule
Average  & 627 & 379 & 67.4 & 4.0 & 324 & 85.5 \\
\bottomrule
\end{tabular}
}
\caption{Details of TPTP-Revised dataset construction. Translation refers to converting the TPTP theorem to Lean 4, while Optimization involves selecting the necessary context.}
\label{tab:tptp-revised-success-rate}
\end{table*}

\section{Computational Efficiency}
\label{sec:efficiency-analysis}

This section provides a detailed analysis of the computational budget and runtime overhead. We evaluate performance under two evaluation paradigms: (1) fixed maximum iteration count, and (2) fixed LLM call budget. 

\subsection{Comparison Under Fixed Iteration Budget}

To ensure a fair comparison with prior work such as Subgoal, which also employs iterative refinement within a bounded number of steps, we set the maximum number of iterations to 10 for all methods. Table~\ref{tab:budget-10iter} reports the average number of LLM calls and token consumption using Claude 3.5, along with proof success rates across benchmark categories. Despite requiring more LLM calls per iteration due to internal branching and feedback mechanisms, DREAM achieves significantly higher overall accuracy (12.7\% vs. 6.7\%).

\begin{table*}[htbp]
\centering
\small
\setlength{\tabcolsep}{4pt}
\renewcommand{\arraystretch}{1.1}
\resizebox{\textwidth}{!}{
\begin{tabular}{lrrrrrrrrrrrrrr}
\toprule
\textbf{Method} & \textbf{Avg. Calls} & \textbf{Avg. Tokens} & \textbf{FLD1} & \textbf{FLD2} & \textbf{GEO6} & \textbf{GEO8} & \textbf{GEO9} & \textbf{GRP5} & \textbf{NUM9} & \textbf{KRS1} & \textbf{SET1} & \textbf{Manual} & \textbf{Avg.} \\
\midrule
Subgoal       & 18  & 53,378 & 5.2\% & 3.1\% & 4.5\% & 4.9\% & 10.0\% & 0.0\% & 7.1\% & 3.0\% & 0.0\% & 29.1\% & 6.7\% \\
DREAM  & 26  & 73,184 & 14.3\% & 12.5\% & 13.6\% & 0.0\% & 0.0\% & 20.0\% & 0.0\% & 3.0\% & 22.2\% & 41.5\% & 12.7\% \\
\bottomrule
\end{tabular}
}
\caption{Computational budget comparison over 10 iterations using Claude 3.5. "Avg. Calls" denotes the average number of LLM calls per solution, and "Avg. Token" represents the average token consumption per solution.}
\label{tab:budget-10iter}
\end{table*}

\begin{table*}[htbp]
\centering
\small
\setlength{\tabcolsep}{5pt}
\renewcommand{\arraystretch}{1.1}
\resizebox{\textwidth}{!}{
\begin{tabular}{lrrrrrrrrrrrrr}
\toprule
\textbf{Method} & \textbf{Max Calls} & \textbf{FLD1} & \textbf{FLD2} & \textbf{GEO6} & \textbf{GEO8} & \textbf{GEO9} & \textbf{GRP5} & \textbf{NUM9} & \textbf{KRS1} & \textbf{SET1} & \textbf{Manual} & \textbf{Avg.} \\
\midrule
Subgoal & 17 & 3.9\% & 3.1\% & 4.5\% & 4.9\% & 10.0\% & 0.0\% & 7.1\% & 3.0\% & 0.0\% & 28.2\% & 6.5\% \\
DREAM   & 17 & 10.4\% & 9.4\% & 13.6\% & 0.0\% & 0.0\% & 20.0\% & 0.0\% & 3.0\% & 22.2\% & 37.4\% & 11.6\% \\
\bottomrule
\end{tabular}
}
\caption{Accuracy comparison under equal LLM call budget (17 calls) using Claude 3.5.}
\label{tab:budget-fixed-calls}
\end{table*}

\subsection{Comparison Under Fixed LLM Call Budget}

To further assess efficiency, we compare DREAM and Subgoal under a constrained total budget of 17 LLM calls—the average number used by Subgoal in the 10-iteration setup. This ensures equivalence in resource usage while isolating algorithmic effectiveness. As shown in Table~\ref{tab:budget-fixed-calls}, DREAM maintains superior performance even under reduced budget, achieving an average accuracy of 11.6\%, compared to 6.5\% for Subgoal. This demonstrates that DREAM’s structured reasoning framework yields higher utility per call.

\section{Quantitative Error Analysis}
\label{sec:error-analysis}

To gain deeper insight into the sources of failure in formal proof generation, we conducted a fine-grained categorization of errors present in unsuccessful proofs produced by DREAM. We employed DeepSeek-V3 as an auxiliary classifier to automatically identify and label error types based on the Lean 4 rejection messages and surrounding proof context. Only error categories occurring more than once are included in the analysis to ensure meaningful interpretation.

\begin{table}[htbp]
\centering
\small
\setlength{\tabcolsep}{10pt}
\renewcommand{\arraystretch}{1.1}
\resizebox{0.4\textwidth}{!}{
\begin{tabular}{lr}
\toprule
\textbf{Error Type}               & \textbf{Frequency} \\
\midrule
type mismatch                     & 151 \\
unknown identifier                & 92  \\
incorrect application             & 26  \\
incorrect tactic                  & 25  \\
unknown tactic                    & 10  \\
unsolved goals                    & 7   \\
unsolved metavariable             & 6   \\
invalid syntax                    & 6   \\
incomplete proof                  & 3   \\
proof strategy error              & 3   \\
declaration conflict              & 2   \\
invalid constructor               & 2   \\
missing dependency                & 2   \\
missing instance                  & 2   \\
\bottomrule
\end{tabular}
}
\caption{Distribution of error types in proofs generated by DREAM using Claude 3.5. Only error types with frequency greater than 1 are displayed.}
\label{tab:error-distribution}
\end{table}

The results, summarized in Table~\ref{tab:error-distribution}, reveal that the majority of errors stem from syntactic and type-level inconsistencies. This highlights a key limitation of LLMs in generating precise formal expressions. Future improvements could therefore benefit from syntax-aware decoding strategies or post-hoc correction modules designed to enforce consistency in types and identifiers.

\section{Dataset Examples}

\begin{figure*}[htbp]
\centering
  \includegraphics[width=\textwidth]{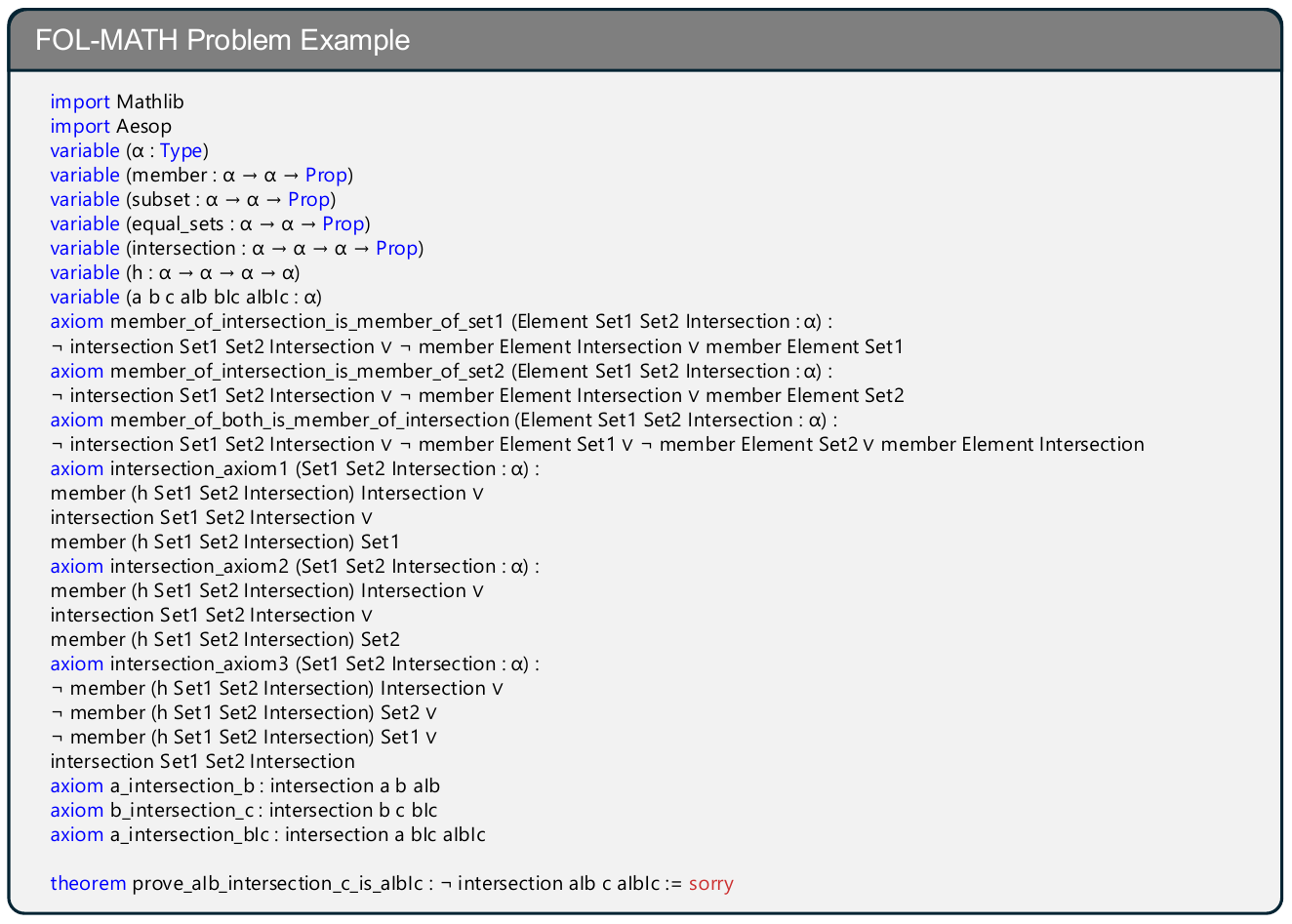}
  \caption{An example of FOL-MATH dataset (SET006).}
  \label{problem-example}
\end{figure*}

An example from our dataset is shown in Table~\ref{problem-example}.

\section{Illustrative Example of the Axiom Tree}
\label{sec:axiom-tree}

\begin{figure*}[htbp]
\centering
  \includegraphics[width=\textwidth]{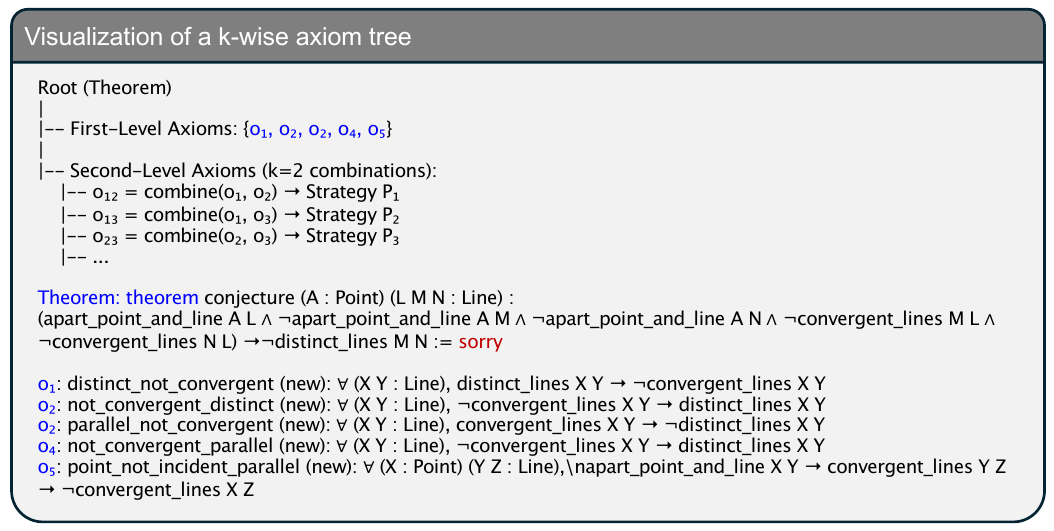}
  \caption{Visualization of a k-wise axiom tree.}
  \label{axiom-tree-visual}
\end{figure*}

We provide a simplified visualization of the hierarchical axiom expansion process used in DREAM (Figure~\ref{axiom-tree-visual}).  Starting from the target theorem, first-level axioms are generated. Subsequently, second-level axioms are derived by combining pairs ($k=2$) of first-level axioms, with each combination yielding a distinct proof strategy.

\section{Examples of Axiom-Driven Strategy Diversification}\label{app:axiom-driven-example}

\begin{figure*}[htbp]
\centering
  \includegraphics[width=\textwidth]{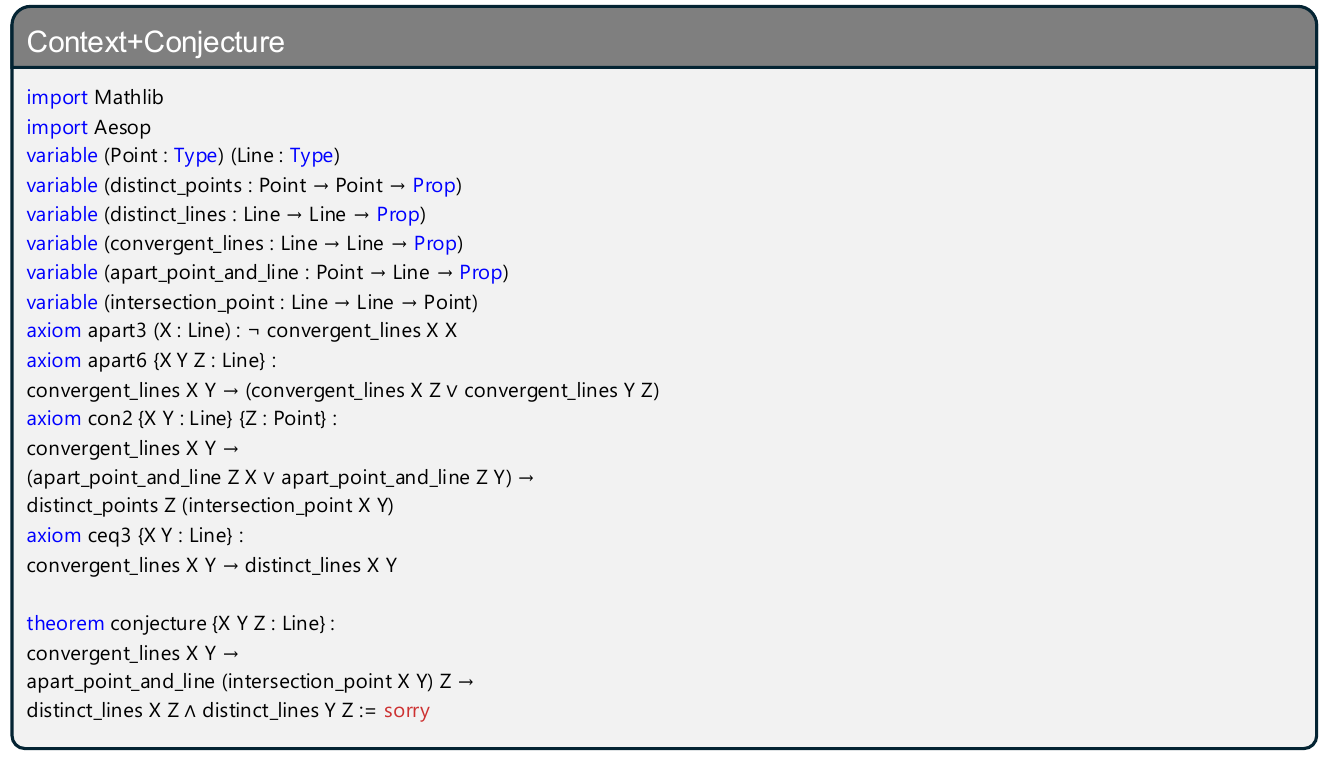}
  \caption{Context and conjecture of the demonstrative example, from which the LLMs generate the following two axiom sets.}
  \label{strategy-example-conjecture}
\end{figure*}

\begin{figure*}[htbp]
\centering
  \includegraphics[width=\textwidth]{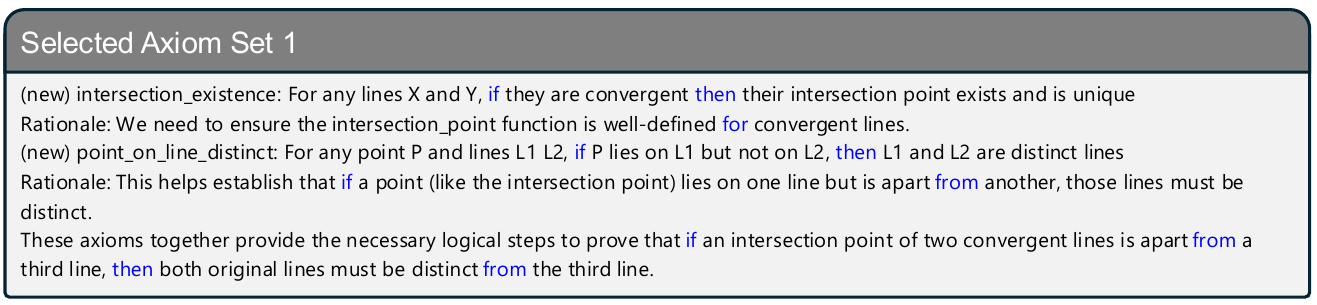}
  \caption{The first set of axioms generated by the LLM based on the given context and conjecture.}
  \label{strategy-example-axiom1}
\end{figure*}

\begin{figure*}[t]
\centering
  \includegraphics[width=\textwidth]{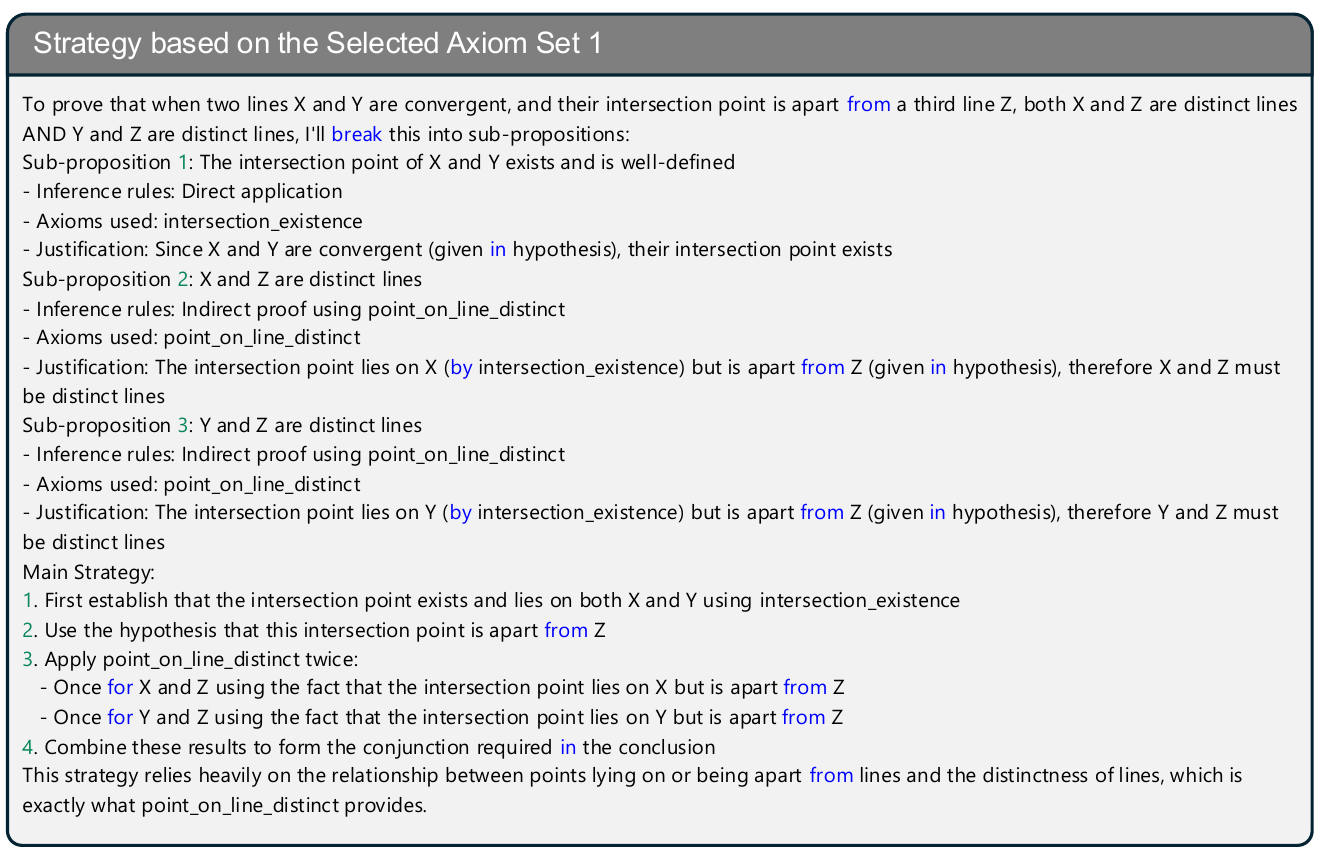}
  \caption{Strategy generated by prompting the LLM to focus on the first set of axioms, utilizing a direct proof method.}
  \label{strategy-example-strategy1}
\end{figure*}

\begin{figure*}[t]
\centering
  \includegraphics[width=\textwidth]{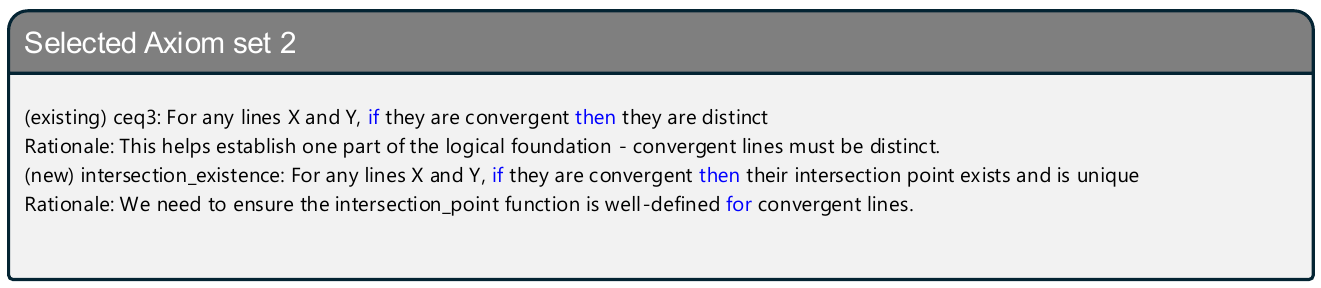}
  \caption{The second set of axioms generated by the LLM based on the given context and conjecture.}
  \label{strategy-example-axiom2}
\end{figure*}

\begin{figure*}[t]
\centering
  \includegraphics[width=\textwidth]{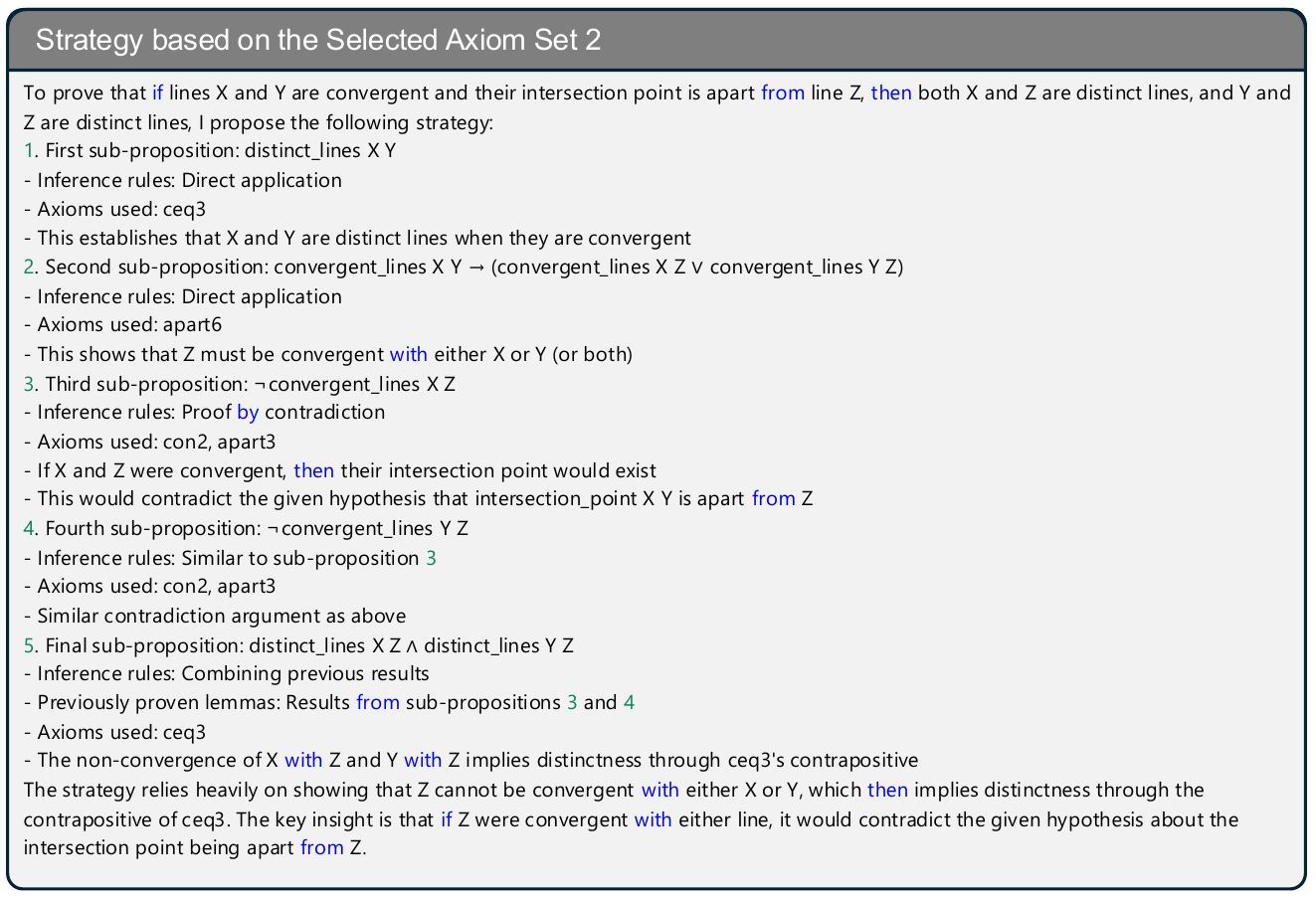}
  \caption{Strategy generated by prompting the LLM to focus on the first set of axioms, employing a proof by contradiction.}
  \label{strategy-example-strategy2}
\end{figure*}

Figures~\ref{strategy-example-conjecture} to \ref{strategy-example-strategy2} illustrate how focusing on different axioms results in varied proving strategies.

\section{Cascading Error in FOL Proofs}\label{app:cascading-error} 

\begin{figure*}[t]
\centering
  \includegraphics[width=\textwidth]{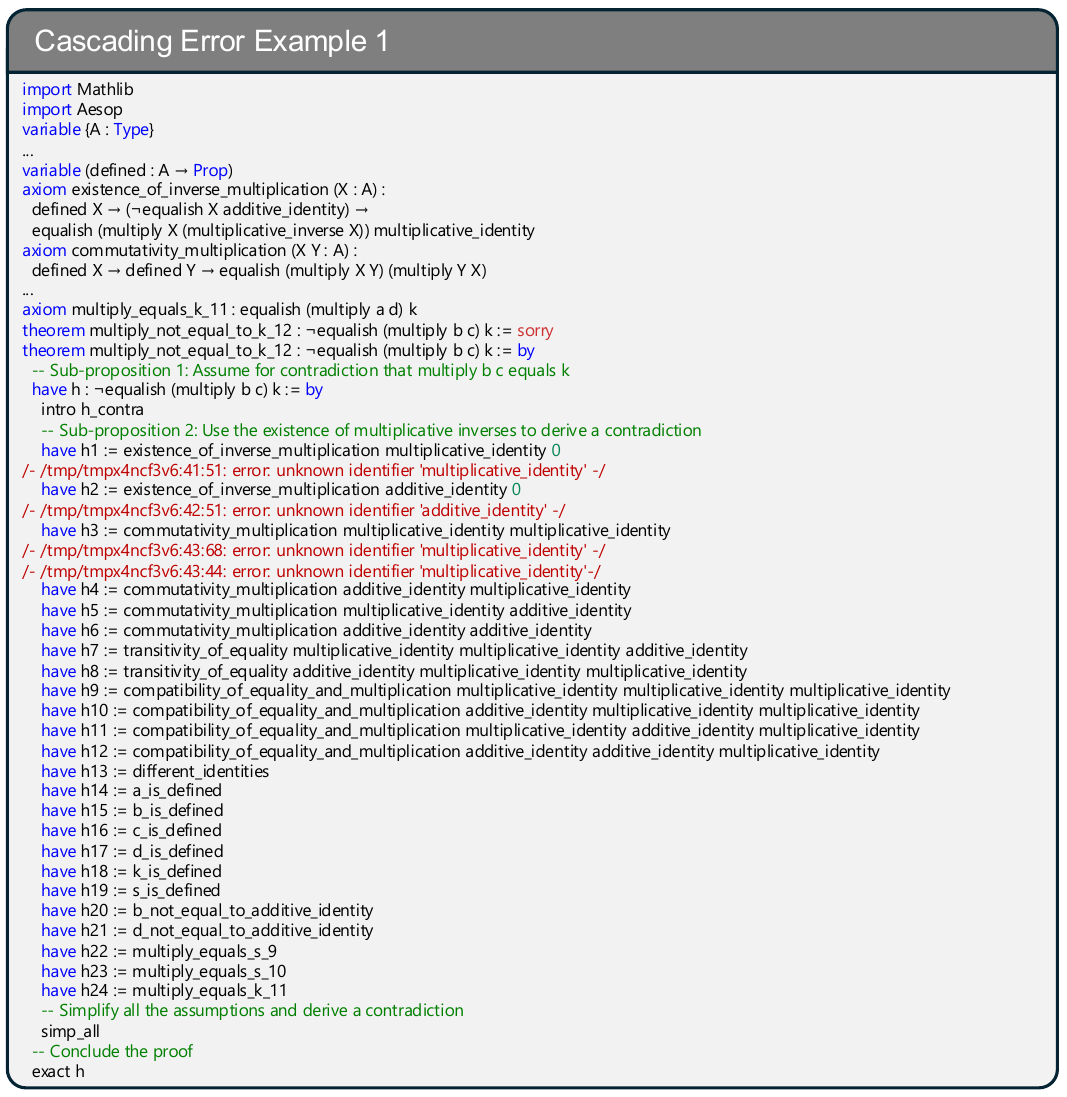}
  \caption{An illustrative example of cascading error.}
  \label{cascading-error}
\end{figure*}

\begin{figure*}[t]
\centering
  \includegraphics[width=\textwidth]{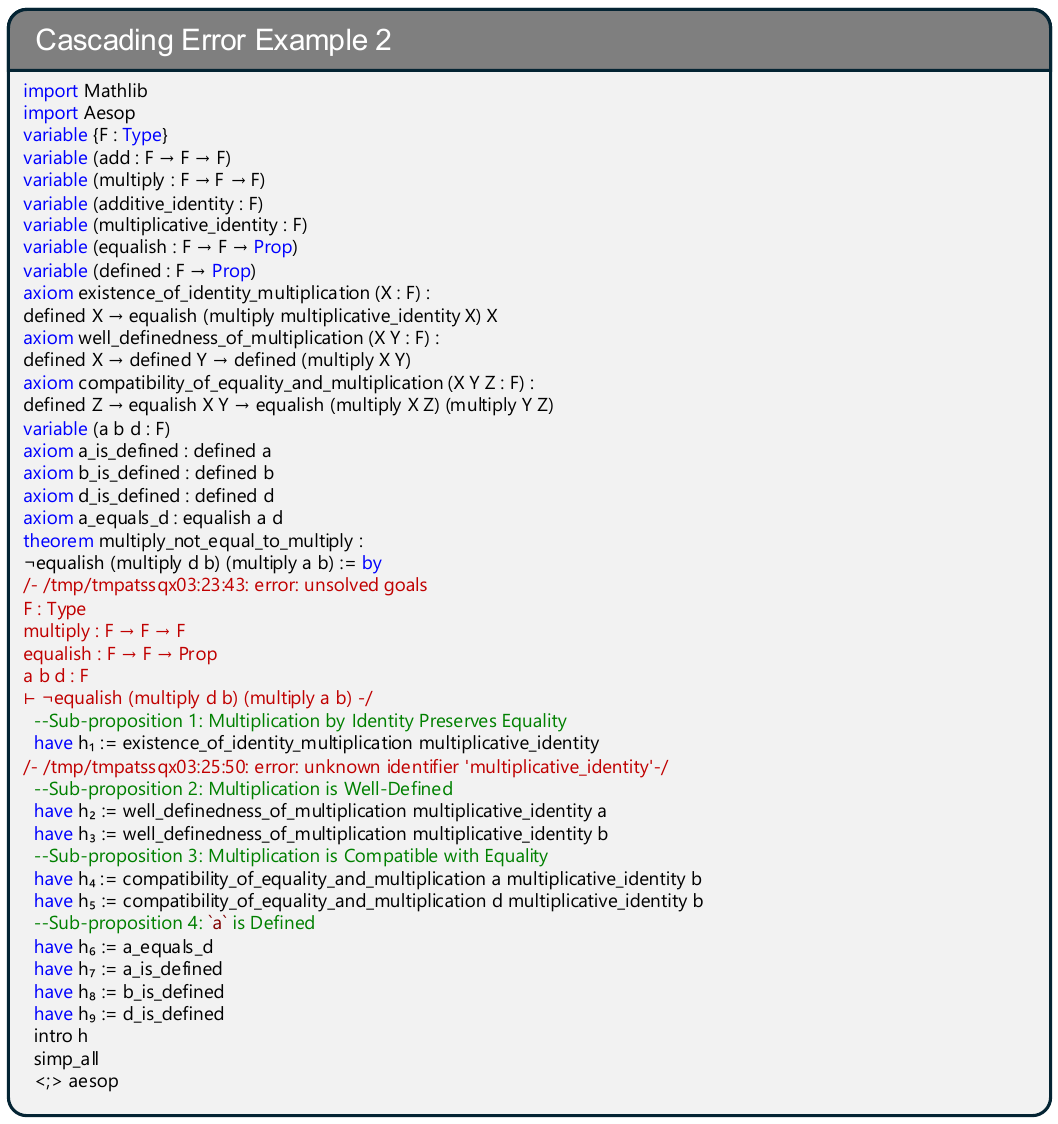}
  \caption{An illustrative example of cascading error.}
  \label{cascading-error-2}
\end{figure*}

Illustrative examples of cascading errors in FOL proofs are presented in Figures~\ref{cascading-error} and \ref{cascading-error-2}.

\section{TPTP to Lean 4 Format Conversion Prompt}
The prompt for converting TPTP to Lean 4 format is shown in Table~\ref{prompt-tptp2lean}.
\begin{figure*}[!ht]
\centering
\begin{lstlisting}
System prompt:
Your task is to convert TPTP format axioms and conjectures into Lean 4 format. Follow these guidelines:

1. Type Declarations:
   - Declare all necessary types using `Type`
   - Define type variables when needed using uppercase letters (e.g., `A`, `B`)

2. Axiom Conversion:
   - Convert each TPTP axiom into a complete Lean 4 definition
   - Use appropriate Lean 4 syntax for logical operators:
   - Do not use `sorry` in axiom definitions

3. Conjecture Conversion:
   - Convert the conjecture into a theorem statement
   - Use `theorem` for the declaration
   - End the theorem with `sorry`
   - Do not provide the proof

4. Code Format:
   - Wrap all Lean 4 code with ```lean``` markers
   - Use proper indentation
   - Include necessary imports
   - Add brief comments explaining complex translations

5. Variable Handling:
   - Declare all variables with appropriate types
   - Maintain consistent variable naming between axioms and conjecture
   - Use meaningful variable names when possible

Please ensure each conversion preserves the original logical meaning while following Lean 4's syntax and type system.

User prompt:
Input TPTP Format:

Axioms:
{axioms}

Conjecture:
{conjecture}

Please provide the Lean 4 conversion following the guidelines above.
\end{lstlisting}
    \caption{Prompts for converting first-order axioms and conjectures from TPTP format to Lean4 format.}
    \label{prompt-tptp2lean}
\end{figure*}
\end{document}